\documentclass[conference]{IEEEtran}
\usepackage{times}
\usepackage{multirow}
\usepackage[numbers]{natbib}
\usepackage{multicol}
\usepackage[bookmarks=true]{hyperref}
\usepackage{graphicx}
\usepackage{color}
\usepackage{array}
\usepackage{algorithm}
\usepackage{algorithmic}
\usepackage{amsmath}
\usepackage{amssymb}
\DeclareMathOperator*{\argmin}{arg\,min}
\usepackage[normalem]{ulem}
\usepackage{float}
\newcolumntype{P}[1]{>{\centering\arraybackslash}p{#1}}

\begin{document}

\title{DexTransfer: Real World Multi-fingered Dexterous Grasping with Minimal Human Demonstrations}


\author{
Zoey Qiuyu Chen$^1$~~~
Karl Van Wyk$^2$~~~
Yu-Wei Chao$^2$~~~
Wei Yang$^2$~~~
Arsalan Mousavian$^2$~~~ \\
Abhishek Gupta$^{1,3}$~~~
Dieter Fox$^{1,2}$~~~
\smallskip 
\\
$^1$University of Washington ~~~
$^2$Nvidia ~~~
$^3$Massachusetts Institute of Technology
}


\makeatletter
\g@addto@macro\@maketitle{
  \begin{figure}[H]
  \setlength{\linewidth}{\textwidth}
  \setlength{\hsize}{\textwidth}
  \centering
  \resizebox{0.98\textwidth}{!}{\includegraphics[]{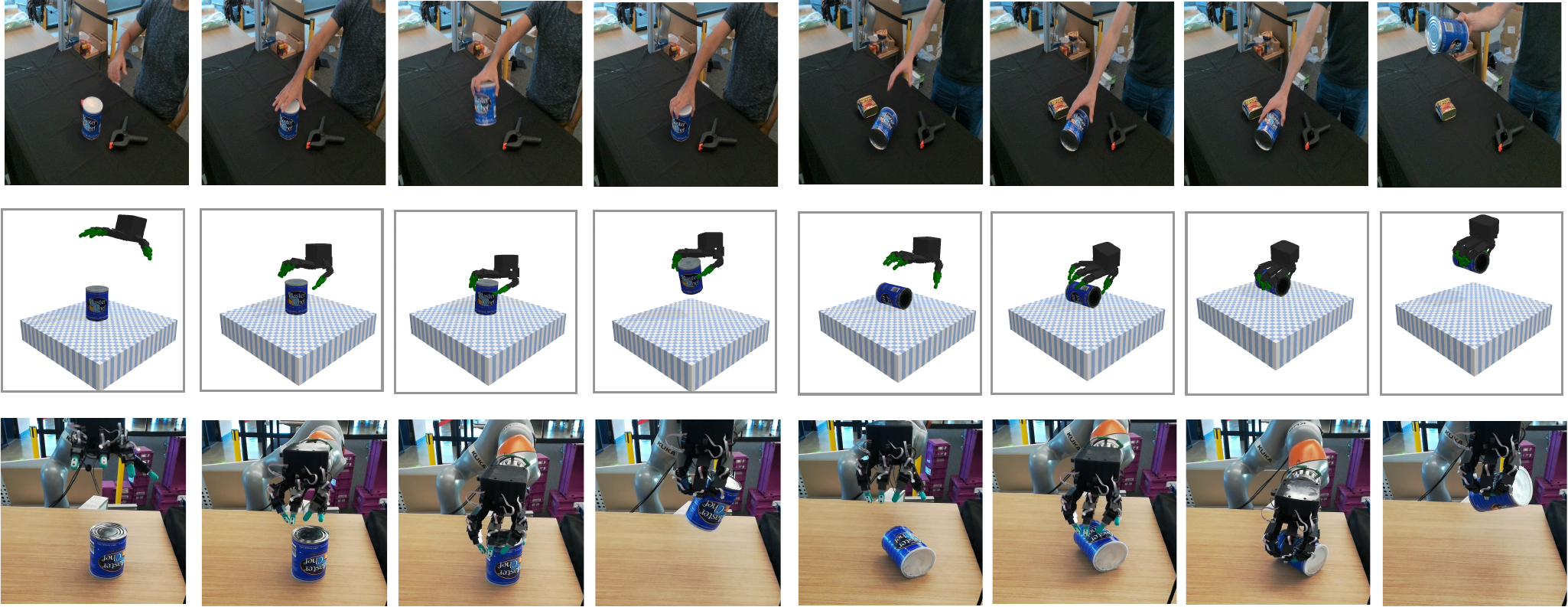}}
  \vspace{-2mm}
  \caption{Illustration of the problem setting for our proposed system \textbf{DexTransfer}. \textbf{Top:} A few human demonstrations are provided via motion capture. \textbf{Middle:} Policies deployed on unseen poses in simulation by training on refined and augmented data based on human demonstrations \textbf{Bottom:} Real Allegro hand executes transferred policy on unseen poses in real world.}
  \label{fig:main_figure}
  \end{figure}
  \vspace{-4mm}
}
\makeatother

\maketitle

\begin{abstract}
Teaching a multi-fingered dexterous robot to grasp objects in the real world has been a challenging problem due to its high dimensional state and action space. We propose a robot-learning system that can take a small number of human demonstrations and learn to grasp unseen object poses given partially occluded observations. Our system leverages a small motion capture dataset and generates a large dataset with diverse and successful trajectories for a multi-fingered robot gripper. By adding domain randomization, we show that our dataset provides robust grasping trajectories that can be transferred to a policy learner. We train a dexterous grasping policy that takes the point clouds of the object as input and predicts continuous actions to grasp objects from different initial robot states. We evaluate the effectiveness of our system on a 22-DoF floating Allegro Hand in simulation and a 23-DoF Allegro robot hand with a KUKA arm in real world. The policy learned from our dataset can generalize well on unseen object poses in both simulation and the real world.

\end{abstract}

\IEEEpeerreviewmaketitle

\section{Introduction}
Our world is largely designed for human hands, anthropomorphic robotic hands are likely to be an important element of robotic systems in human-centric environments. 
To build general purpose dexterous manipulation systems, we require methods that are able to train policies that can generalize widely with relatively little burden placed on a human supervisor. 
In this work, our goal is to build a system that is able to leverage a small amount of human supervision to learn a robust controller for dexterous manipulation tasks, which can be reliably deployed on a real robot.

We proposed a demonstration guided data augmentation system (Fig.\ref{fig:overview}) that aims to generate a large dataset of diverse, successful trajectories for a robot gripper in simulation. Our data-augmentation pipeline combines motion retargeting with local gradient-free trajectory refinement and augmentation to obtain a large variety of successful data relatively cheaply. This data can be used to learn policies mapping from point-clouds to actions, which can be transferred to real world scenarios to grasp objects in novel poses. 

While human demonstration plays a crucial role in training manipulation policies, devising a scalable method to collect demonstrations is still a key bottleneck. Prior work has collected demonstrations kinesthetically~\cite{zhu:icra2019}, through VR interfaces~\cite{rajeswaran:rss2018}, or using motion capture (mocap) solutions~\cite{garcia-hernando:cvpr2018,hasson:cvpr2019,taheri:eccv2020,qin:arxiv2021}. Some recent efforts have also explored collecting demonstrations by harvesting internet video~\cite{mandikal:corl2021}. Although there has been some innovations for parallel-jaw grippers~\cite{kokic:ral2020,song:ral2020,young:corl2020}, acquiring demonstrations for multi-fingered dexterous manipulators remains challenging. Consequently, the collected demonstration datasets are often limited in size. Our work leverages existing mocap datasets of human grasping~\cite{chao:cvpr2021} to guide the generation of a large, diverse grasping dataset for training dexterous manipulators.

\section{DexTransfer: A System for Real World Dexterous Grasping with Minimal Human Demonstrations}
\label{sec:method}
\setcounter{figure}{1}
\begin{figure}[!ht]
    \centering
    \includegraphics[width=0.49\textwidth]{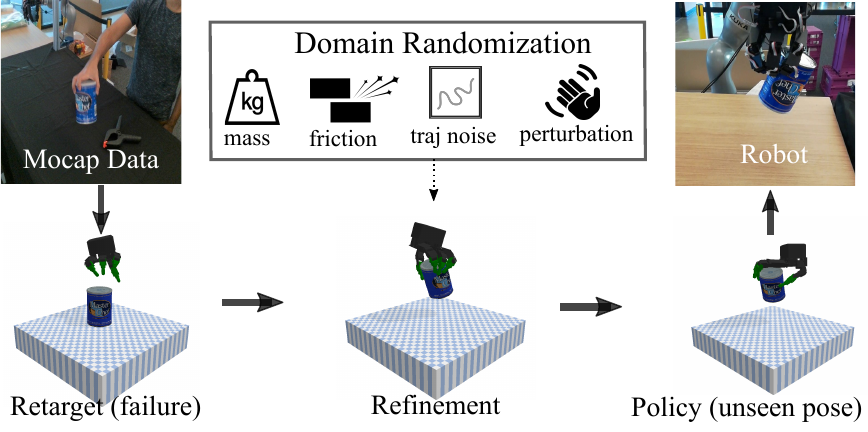}
    \caption{Overview of the proposed framework DexTransfer. The mocap data is first retargeted to a dexterous gripper in simulation. This motion reference is then refined and augmented into a large and diverse set of successful trajectories, to learn a policy to succeed on unseen object poses and initial hand poses. The learned policy is eventually transferred to a real robot system.}
    \label{fig:overview}
\end{figure}

Our work aims to build a system that can leverage a small number of human demonstrations to learn grasping policies for dexterous grippers which are robust across a variety of object poses in the real world. To accomplish this, our system consists of three major phases - (1) trajectory retargeting (2) trajectory refinement and (3) policy learning, that are performed in sequence to acquire dexterous grasping policies as illustrated in Figure~\ref{fig:overview}. The system starts by being provided with human demonstrations, which are first retargeted to an actual robot gripper in simulation as described in Section~\ref{sec:retarget}. Since the retargeted trajectories are not guaranteed to be dynamically successful, the system then aims to generate trajectory refinements that leads to robust and successful grasps as described in Section~\ref{sec:trajectory_refinement}, along with performing directed trajectory augmentation to increase the dataset size and diversity. Lastly, these successful grasp trajectories can be used for policy learning directly from sensory observations in Section~\ref{sec:policy_learning}.

\subsection{Trajectory Retargeting}
\label{sec:retarget}

Our system takes a set of human demonstrations provided via a hand motion capture system. Precisely, we assume a demonstration dataset $\mathcal{D}_{\text{human}} = \{\tau_0^h, \tau_1^h, \dots, \tau_N^h\}$, where each demonstration $\tau_i = \{(q_0^i, o_0^i), (q_1^i, o_1^i), \dots, (q_T^i, o_T^i)\}$ is a trajectory of hand 3D pose $q_t$ and object 3D pose $o_t$. These demonstrations, when directly mapped to a robot hand using standard Inverse Kinematics \cite{gleicher:siggraph1998}, are typically not functional due to the mismatch between the human and robot actuator shape and kinematics. 

To address this issue, we follow DexPilot~\cite{handa:icra2020} and formulate the retargeting objective as a non-linear optimization problem with a general cost function that is able to retarget human data to various robot hands of different morphologies while preserving the original demonstration behaviors.

\begin{align}
    d_f & = \sum_{i=0}^{N} \left\| r_i(\mathbf{q}_r) - s_r \hat{r}_i(\mathbf{q}_h) \right\|^2, \label{eq:df}\\
    d_{obj} &= \left\| o(\mathbf{f}_r) - o(\mathbf{f}_h) \right\|_2, \label{eq:dobj}\\
    d_r & = \mathbf{G}(\mathbf{M}_r, \mathbf{M}_h), \label{eq:dr}\\
   \argmin_{(\mathbf{q}_r, \mathbf{f}_r, \mathbf{M}_r)} & (w_f d_f + w_{obj} d_{obj} + w_r d_r),
\end{align}

Eq.~\ref{eq:df} is the same as the cost function defined in \cite{handa:icra2020}, where each $r_i(\mathbf{q}_r)$ and $\hat{r}_i(\mathbf{q}_h)$ are displacement vectors between hand joints for robot and human hands, and $s_r$ is the scale ratio between allegro and a human hand. To move the robot hand towards the object in a similar way as the demonstration, we introduce Eq.~\ref{eq:dobj} where $o(\mathbf{f}_r)$ and $o(\mathbf{f}_h)$ represents displacements between the object center and each finger tip. We also add Eq.~\ref{eq:dr} where $\mathbf{G}$ is the minimum geodesic distance in SO(3) between the orientation of the human palm $\mathbf{M}_h$ and robot palm $\mathbf{M}_r$.

The output of retargeting is a set of demonstrations represented by the joint positions of the robot $\mathcal{D}_{\text{retargeted}} = \{\tau_0^r, \tau_1^r, \dots, \tau_N^r\}$, with each trajectory $\tau_i^r = \{(h_0^i, e_0^i, o_0^i), (h_1^i, e_1^i, o_1^i),, \dots, (h_T^i, e_T^i, o_T^i),\}$ consisting of finger positions $h_t^i$, end effector positions $e_t^i$ and object poses/pointclouds $o_t^i$. These roughly capture human trajectories while being consistent with the robot kinematics. We parameterize the action space as target joint positions and we run an underlying PID controller, making it trivial to define actions $a_t^i$ from these demonstrations simply as the next position in the trajectory $(h_{t+1}^i, e_{t+1}^i)$.

\subsection{Trajectory Refinement}
\label{sec:trajectory_refinement}
Retargeted trajectories simply try to match kinematic poses without actually considering the dynamics of the world, not accounting for unseen contact forces. For contact rich manipulation tasks like dexterous grasping, this can lead to catastrophic failures in trajectory execution. The process of trajectory refinement takes the (potentially unsuccessful) retargeted trajectories $\mathcal{D}_{\text{retargeted}} = \{\tau_0^r, \tau_1^r, \dots, \tau_N^r\}$ from Section~\ref{sec:retarget} and refines them to a diverse set of successful trajectories on a variety of different objects. The process of trajectory refinement consists of (1) Generating a large set of nominal trajectories to refine from retargeted trajectories via template matching (2) perturbing and refining these nominal trajectories to be dynamically successful at task completion (3) augmenting refined trajectories to be diverse across poses, configurations and initial hand states. We describe each of these components in detail. 

\textbf{Template Matching for Nominal Trajectory Generation}
Typically we are only given a handful of demonstrations by a human supervisor. For data driven policy learning methods as described in Section~\ref{sec:policy_learning}, this is hardly sufficient and we need to generate a significantly larger dataset to learn general purpose policies. The first insight we leverage is that for rigid bodies, when we apply a transformation $\mathcal{T}$ of the object pose in SE(3) space, the same transformation $\mathcal{T}$ can also be applied to the end-effector trajectories $e_t^i$ (in most cases) to yield sensible trajectories $\mathcal{T}e_t^i$ for the new object pose as well. For each initial object pose $o_0^i$ that is encountered in a particular demonstration $\tau_i$, we can estimate the rigid transform $\mathcal{T}_{o_0^i}^{o_0^j}$ to the object pose $o_0^j$ in a different demonstration $\tau_j$ and use it to transform the first demonstrations end effector trajectory $e_t^i$ to the second objects frame of reference $\mathcal{T}_{o_0^i}^{o_0^j}e_t^i$ and vice versa. More precisely,

\begin{align*}
   o_0^j =& \mathcal{T}_{o_0^i}^{o_0^j}e_t^i, \hspace{0.3cm}\forall i, j \\ 
   \hat{\tau}_j = \{(h_0^j, \mathcal{T}_{o_0^i}^{o_0^j}e_0^i, o_0^j), (h_1^j, &\mathcal{T}_{o_0^i}^{o_0^j}e_1^i, o_1^j), \dots, (h_T^j, \mathcal{T}_{o_0^i}^{o_0^j}e_T^i, o_T^j)\}, \nonumber\\
\end{align*}

This transformation provides a drastically larger number of effective trajectories performing in a variety of different object poses \footnote{In our work, we restrict the transformations to be between just a set of stable canonical poses and perform after the fact rotations to further increase diversity. We compute stable canonical poses using trimesh}. Note that these trajectories do not actually have to be perfect or successful when executed in the environment as they are refined in the following phase of the pipeline, but the large diversity of such ``nominal" trajectories helps with policy learning in Section~\ref{sec:policy_learning}. We denote the trajectories obtained after the template matching phase as $\mathcal{D}_{\text{template}}$

\textbf{Refinement via Correlated Sampling}
Retargeted trajectories are not successful at dynamically solving the task as they simply match kinematics but not the dynamics of the desired behavior (including desired contact forces). We found that directly replaying the actions in a trajectory open-loop, typically results in close to zero successes. This suggests that to generate an appropriate dataset for supervised learning, we need to refine the original retargeted trajectories to be dynamically successful in simulation. 

Our key insight here is that a simple technique based on perturbation with rejection sampling can be effective in generating refined trajectories retargeted from demonstrations. When considering how to refine trajectories, we perturb the motion in the neighborhood of retargeted trajectories so as to find dynamically successfully trajectories that grasp and lift the object. The most naive approach to this would be to randomly add independent gaussian noise to various actuators to create perturbations around the nominal trajectory so as to find a more successful set of controls. However, in a high dimensional state space such as multi-fingered dexterous grasping, this quickly becomes ineffective. Instead, we perform perturbations in a more directed ``synergy" space which allows for coordinated perturbations of several fingers so as to open and close fingers in a coordinated fashion, rather than simply perturbing joints independently.

\begin{algorithm}[!h]
\label{algo:refinement}
\caption{Correlated Sampling for Refinement}
\textbf{Input:} Retargeted trajectories $\mathcal{D}_{\text{template}}$  \\
\textbf{Output:} Refined trajectories $\mathcal{D}_{\text{refined}}$

\begin{algorithmic}
\STATE Initialize $\mathcal{D}_{\text{refined}} = \{\emptyset\}$
\WHILE{	$\lvert\mathcal{D}_{\text{refined}}\rvert \leq. N$} 
    \STATE Sample $t \sim \mathcal{U}[0, 1]$
    \STATE Compute perturbation $(p_t^i)_k = t*(p_{\text{max}}^i)_k + (1-t)*(p_{\text{min}}^i)_k \forall t, i, k$
    \STATE Compute perturbed action $\hat{(a_t^i)}_k = (a_t^i)_k + (p_t^i)_k$
    \STATE Execute open loop action sequence $\{\hat{(a_0^i)}, \hat{(a_1^i)}, \dots, \hat{(a_T^i)}\}$ to get trajectory $\hat{\tau_i}$
    \STATE If $\hat{\tau_i}$ passes stability checks, $\mathcal{D}_{\text{refined}} = \mathcal{D}_{\text{refined}} \bigcup \hat{\tau_i}$
\ENDWHILE 
\STATE{\textbf{return} $\mathcal{D}_{\text{refined}}$}
\end{algorithmic}
\end{algorithm}

To be more precise, at every step of the trajectory the true control $a_t^i$ of a retargeted trajectory ($\tau_i$) is perturbed by sampling and applying correlated perturbations to the nominal controls of the retargeted trajectory. These correlated perturbations are sampled by first sampling a scalar parameter time $t$ from the uniform distribution $t \sim \mathcal{U}[0, 1]$ and then using this parameter to choose a per-joint  perturbation $(p_t^i)_k = t*(p_{\text{max}}^i)_k + (1-t)*(p_{\text{min}}^i)_k$ that interpolates between minimum and maximum perturbation values per joint $k$, which can then be added to $a_t^i$ to obtained the perturbed action. This sampling scheme allows for correlated perturbation of the fingers, allowing for perturbation to ``open" and ``close" the fingers more coherently, rather than simply doing uniformly random exploration independently per-joint, which makes refinement significantly less effective. 

To generate successfully refined trajectories, we can generate perturbations in the control as described above, simulate trajectories via standard forward simulation and then reject trajectories which don't meet the desired stability (object does not fall under perturbation and randomization of parameters like mass and friction and is lifted high off the table) and success criteria (object is grasped and lifted). Our choice of rejection condition during rejection sampling can allow us to select for the most robust behaviors as described in Algorithm~\ref{algo:refinement}.

\textbf{Object-Centric Augmentation.}
While refinement yields dynamically consistent trajectories $\mathcal{D}_{\text{refined}}$ that successfully apply contact forces to grasp objects, it still only covers a somewhat narrow diversity of object, hand and arm poses since it performs very local refinements on nominal trajectories. Given that most household objects have multiple plausible grasp positions, which simply differ by a translation offset, we can perform a simple \emph{augmentation} scheme to propose random translation offsets to refined trajectories and retain those translationally perturbed trajectories that are still able to succesfully accomplish the grasping task. This allows us to synthesize a larger dataset of different grasping trajectories beyond the human demonstrations. 

While refinement and augmentation do generate a variety of different grasping trajectories, the number of distinct initial positions of the end effector is still somewhat minimal. This makes policy generalization to novel end effector positions challenging. Given the nature of the grasping problem, we can synthetically generate a much larger dataset $\mathcal{D}_{\text{augmented}}$ of trajectories with varying initial hand poses by generating a variety of initial hand positions and then doing interpolation in free space to the closest trajectory amongst the existing human provided demonstrations, from where that demonstration can subsequently be executed. We refer this step as data funneling. 

Given the combination of template matching, refinement with correlated sampling and object-centric augmentation with funneling, we finally arrive at a diverse and dynamically successful dataset $\mathcal{D}_{\text{augmented}}$ that is able to successfully apply contact forces and perform grasping from a variety of configurations. This is then used for policy learning as described below. 

\begin{table*}
\centering
\caption{Allegro Hand Dataset across 17 YCB objects. We transfer a small set of Mocap Data to a large dataset with successful trajectories.}
\label{table:dataset}
\begin{tabular}{|P{3cm}|P{2cm}|P{2cm}|P{2cm}|P{2cm}|P{2cm}|P{2cm}|} 
\hline
\begin{tabular}[c]{@{}l@{}}\\\end{tabular} & \underline{master chef can}      & \underline{~ ~cracker box~ ~}   & \underline{~ ~ ~sugar box~ ~ ~} & \underline{tomato soup can}    & \underline{~ mustard bottle~}  & \underline{~ ~tuna fish can~~}             \\
Human Demo                                 & 26                           & 25                          & 25                          & 25                         & 25                         & 28                                     \\
Normalized Trajectory                      & 78                           & 141                         & 158                         & 56                         & 75                         & 122                                    \\

Refined Instance                           & 182                          & 100                         & 313                         & 146                        & 82                         & 200                                    \\
\textbf{Augmented~Trajectory}              & \textbf{286}                 & \textbf{462}                & \textbf{1018}               & \textbf{294}               & \textbf{46}                & \textbf{139}                           \\ 
\hline
                                          & \underline{~ ~pudding box~ ~}    & \underline{~ ~ gelatin box~ ~~} & \underline{potted meat can}     & \underline{~ ~wood block~~}    & \underline{~ ~foam brick~~}    & \underline{~ ~ ~ ~bowl~ ~ ~ ~}             \\
Human Demo                                 & 23                           & 27                          & 23                          & 26                         & 26                         & 27                                     \\
Normalized Trajectory                      & 87                           & 118                         & 128                         & 102                        & 104                        & 79                                     \\
Refined Trajectory                         & 40                           & 52                          & 96                          & 55                         & 79                         & 25                                     \\

\textbf{Augmented Trajectory}              & \textbf{110}                 & \textbf{248}                & \textbf{713}                & \textbf{846}               & \textbf{380}               & \textbf{179}                           \\ 
\hline
                                          & \underline{~ ~ ~ ~ mug~ ~ ~ ~ ~} & \underline{~bleach cleaner~}    & \underline{~ ~power drill~ ~}   & \underline{~ ~ large marker~~} & \underline{~ ~ pitcher base~~} & \underline{~ ~ ~ ~\textbf{sum}~ ~ ~ ~}  \\
Human Demo                                 & 23                           & 23                          & 23                          & 26                         & 25                         & 426                                    \\
Normalized Trajectory                      & 101                          & 56                          & 64                          & 78                         & 74                         & 1621                                   \\

Refined Instance                           & 122                          & 102                         & 45                          & 43                         & 34                         & 2566                                   \\
\textbf{Augmented~Trajectory}              & \textbf{109}                 & \textbf{181}                & \textbf{123}                & \textbf{9}                 & \textbf{38}                & \textbf{5181}                          \\
\hline
\end{tabular}
\end{table*}

\subsection{Policy Learning from Point Cloud observations}
\label{sec:policy_learning}
\begin{figure}[h]
    \centering
    \includegraphics[width=0.49\textwidth]{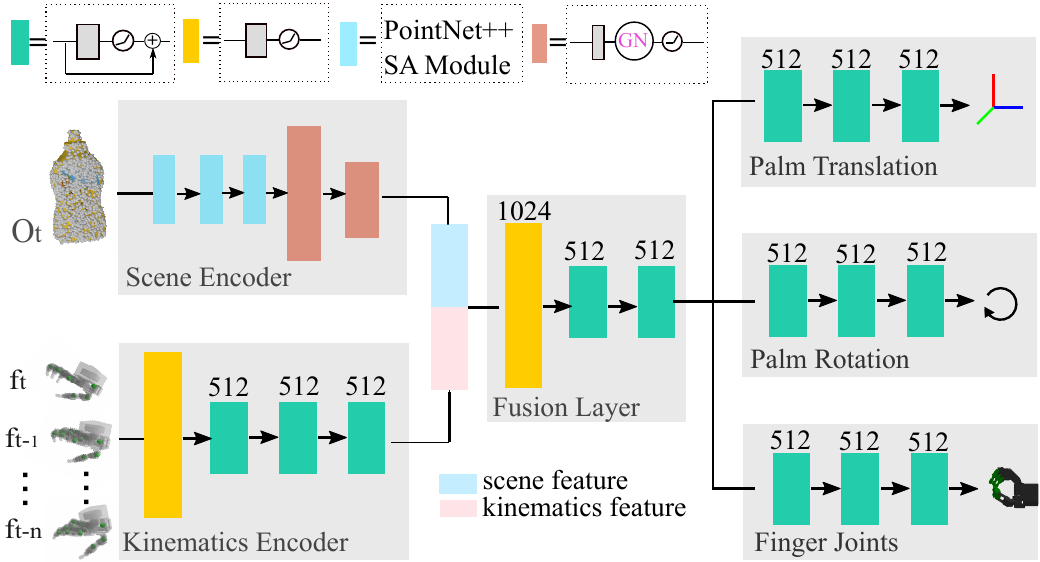}
    \caption{Network architecture for the point cloud dependent policy $\pi_{\theta}$. \textbf{Scene Encoder} consists of three PointNet++ SA modules followed by two fully-connected layers. \textbf{Kinematics Encoder} consists of three residual modules. \textbf{Fusion Layer} takes the concatenated features and feed into one linear layer followed by two residual modules. The network has three branches to predict palm translation, rotation and joint angles. Each branch consists of three residual modules.}
    \label{fig:network_arch}
\end{figure}

Given a robust and diverse dataset of robot trajectories, we train a policy to directly predict the action to execute based on current observation via a standard maximum likelihood supervised learning objective. Our network consists of a \textbf{Scene Encoder} and a \textbf{Kinematics Encoder}. The Scene Encoder is based on PointNet++ \cite{qi:nips2017}, that takes point clouds of the object as current observation $O_t$, and predicts scene features as output. The Kinematics Encoder takes the past 5 frames of motions (to deal with partial observability and variable timing) ${f_t, f_{t-1}...f_{t-4}}$, each includes robot finger joints, keypoints and object center shift, and outputs the kinematics feature. The scene features and kinematics features are then concatenated and fed into Fusion Layer to compute combined features for the current observation. The network splits three branches to predict translation (3-dim), rotations (3-dim) and finger joints (N-dim, N=16 for Allegro). We found that representing observation and action relative to robot palm coordinate is crucial for generalizing to unseen object poses. Thus all the observation and keypoint positions are represented in the current palm coordinates. The predictions of the network are used to compute the loss function as follows, which can then be used to learn policy parameters $\theta$ via standard stochastic gradient methods: 

\begin{align}
   \min_{\theta} \mathbb{E}_{(h_t^i, e_t^i, o_t^i, a_t^i) \sim \mathcal{D}_{\text{aug}}}& \|\pi_{\text{translation}}^\theta(\cdot \mid h_t^i, e_t^i, o_t^i)  - (a_t^i)_{\text{translation}}\| \\ & +  \mathbf{G}(\pi_{\text{rotations}}^\theta(\cdot \mid h_t^i, e_t^i, o_t^i), (a_t^i)_{\text{rotations}}) \\ & + \|\pi_{\text{finger}}^\theta(\cdot \mid h_t^i, e_t^i, o_t^i) - (a_t^i)_{\text{finger}}\| 
\end{align}

where $\mathbf{G}$ is the minimum geodesic distance in SO(3).

\section{Experiments}
In this section, we aim to answer the following questions: 
\begin{enumerate}
    \item Can DexTransfer enable us to leverage a small number of human demonstrations to obtain a large number of dynamically successful trajectories in sim across a variety of objects?
    \item Can we understand which elements of the proposed system are most crucial to enabling robust dexterous grasping performance?
    \item Do the resulting policies transfer to the real world on an Allegro robot hand?
\end{enumerate}

We show videos experiments for both simulation and real world in the supplementary material and further results can be found on the supplementary website \mbox{\url{https://sites.google.com/view/dextransfer/home}}

\subsection{Dataset Generation in Simulation}
\label{sec:dataset}
We extracted human demonstrations from the DexYCB dataset \cite{chao:cvpr2021}, a dataset with 1,000 mocap sequences of humans picking up YCB objects \cite{calli:icar2015} from a table. Each sequence contains a single human subject picking up one specific object with a single hand. The hand pose is captured using the MANO hand model~\cite{romero:siggraphasia2017}. Given the raw dataset, we curated a subset with 17 objects picked up by a human with the right hand, resulting in a total of $426$ demonstration sequences with varying object configurations. Table \ref{table:dataset} reports the accumulated number of trajectories after each processing stage for the Allegro hand in a per object breakdown. Despite MoCap tracking error and kinematics difference between the robot and the human demonstrator, DexTransfer is able to generate more than 10 folds successful data for the allegro robot. 

\begin{figure*}[h]
    \centering
    \includegraphics[width=0.8\textwidth]{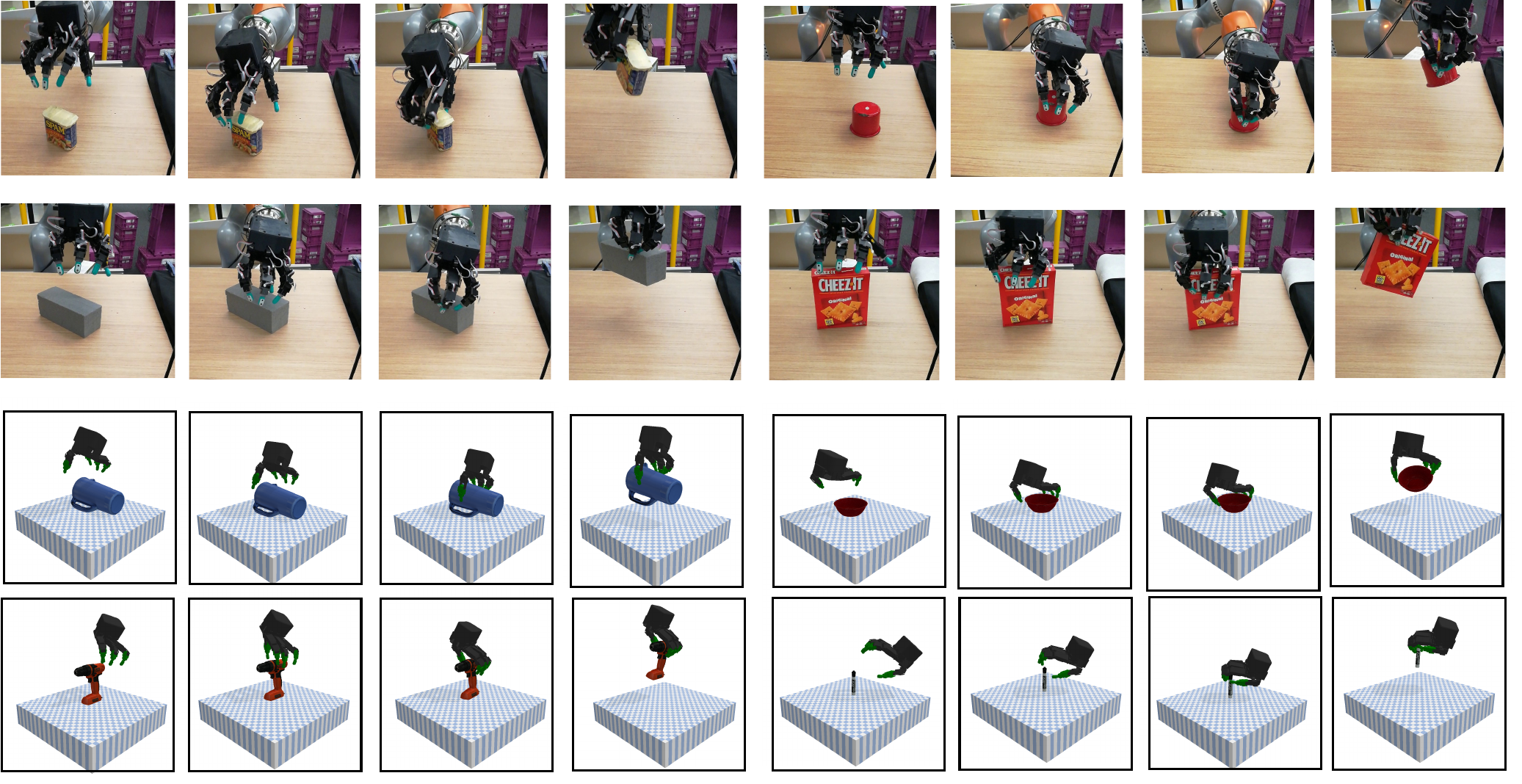}
    \caption{Qualitative results of policies on unseen poses of various objects in both simulation and real world. We can see that the hand approaches the objects from a variety of approach angles and grasp poses and can show interesting grasping strategies.}
    \label{fig:combined_qualitative}
\end{figure*}

\subsection{Policy Learning on Augmented Dataset}
In this section, we show how we can leverage the augmented dataset to learn policies that can perform well at dexterously grasping objects in unseen poses and configurations. In particular, we follow the procedure described in Section~\ref{sec:policy_learning} and learn a per-object policy that is able to dexterously grasp a particular object in a variety of different poses. During test time, we randomize object 6D poses and initial positions of the robot. We evaluate our policies 20 times on each object. Similar to the set up in \cite{sundermeyer2021contact}, each time, if the robot drops the object, we reset our network to additionally captures the rate of success within three attempts. Note that a test episode is ended as soon as a success is reached. The success is defined when the object is above the table by 10 cm. In Table \ref{table:baselines}, we show the success rate of the robot after each attempt. The policy is able to recover and succeed in the next trials because its trained in a closed-loop fashion. All experiments are trained with the same batch size $bs=460$ and Adam optimizer with learning rate $lr = 0.0002$. We use Nvidia Apex \cite{apex} to achieve mixed-precision training through all of our experiments. For depth rendering, we use pyrender \cite{pyrender} to enable fast online rendering during training. We show the policy is able to grasp most objects with a success rate of over $70\%$ on unseen poses and configurations in Table \ref{table:baselines}(DexTransfer (ours)).

Overall, these results indicate that the augmented dataset combined with policy learning via supervised learning allow us to acquire dexterous grasping behavior that is general both across unseen poses and across different objects. We additionally show qualitative results of Allegro hand grasping objects with different poses at different frames in Fig.\ref{fig:combined_qualitative}.

\subsection{Understanding Different Elements of the Refinement Pipeline}
\begin{table*}
\centering
\caption{Ablation Study in Simulation}
\label{table:baselines}
\begin{tabular}{|l|l|l|l|l|l|l|} 
\hline
                           & \uline{master chef can}      & \uline{~ ~cracker box~ ~}   & \uline{~ ~ ~sugar box~ ~ ~} & \uline{tomato soup can}    & \uline{~ mustard bottle~}  & \uline{~ ~tuna fish can~~}                      \\
                           & ~ 1~ ~ ~2~ ~ ~3              & ~ 1~ ~ ~2~ ~ ~3             & ~ 1~ ~ ~2~ ~ ~3             & ~ 1~ ~ ~2~ ~ ~3            & ~ 1~ ~ ~2~ ~ ~3            & ~ 1~ ~ ~2~ ~ ~3                                 \\
Heuristic                  & 0.00~ 0.00~ 0.00             & 0.00~ 0.00~ 0.00            & 0.25~ 0.30~ 0.35            & 0.05~ 0.05~ 0.05           & 0.00~ 0.00~ 0.00           & 0.20~ 0.20~ 0.20                                \\
Nearest Neighbor           & 0.20~ 0.20~ 0.20             & 0.10~ 0.10~ 0.10            & 0.10~ 0.10~ 0.10            & 0.10~ 0.10~ 0.10           & 0.25~ 0.25~ 0.25           & 0.25~ 0.25~ 0.25                                \\
No Funneling               & 0.05~ 0.05~ 0.05             & 0.05~ 0.05~ 0.05            & 0.10~ 0.15~ 0.15            & 0.00~ 0.00~ 0.00           & 0.05~ 0.05~ 0.05           & 0.05~ 0.10~ 0.10                                \\
No Randomization           & 0.65~ 0.80~ 0.80             & 0.25~ 0.25~ 0.25            & 0.35~ 0.35~ 0.35            & 0.25~ 0.25~ 0.25           & 0.50~ 0.55~ 0.55           & 0.40~ 0.55~ 0.55                                \\
No Augmentation            & 0.65~ 0.65~ 0.65             & 0.20~ 0.20~ 0.20            & 0.60~ 0.65~ 0.65            & 0.55~ 0.60~ 0.60           & 0.50~ 0.50~ 0.50           & 0.40~ 0.50~ 0.55                                \\
\textbf{DexTransfer(ours)} & \textbf{0.70~ 0.80~ 0.80}    & \textbf{0.25~ 0.30~ 0.35}   & \textbf{0.65~ 0.70~ 0.80}   & \textbf{0.70~ 0.85~ 0.85}  & \textbf{0.75~ 0.85~ 0.85}  & \textbf{0.70~ 0.80~ 0.85}                       \\ 
\hline
                           & \uline{~ ~pudding box~ ~}    & \uline{~ ~ gelatin box~ ~~} & \uline{potted meat can}     & \uline{~ ~wood block~~}    & \uline{~ ~foam brick~~}    & \uline{~ ~ ~ ~bowl~ ~ ~ ~}                      \\
                           & ~ 1~ ~ ~2~ ~ ~3              & ~ 1~ ~ ~2~ ~ ~3             & ~ 1~ ~ ~2~ ~ ~3             & ~ 1~ ~ ~2~ ~ ~3            & ~ 1~ ~ ~2~ ~ ~3            & ~ 1~ ~ ~2~ ~ ~3                                 \\
Heuristic                  & 0.20~ 0.20~ 0.20             & 0.05~ 0.05~ 0.05            & 0.30~ 0.30~ 0.30            & 0.00~ 0.00~ 0.05           & 0.15~ 0.15~ 0.15           & 0.00~ 0.00~ 0.00                                \\
Nearest Neighbor           & 0.20~ 0.25~ 0.25             & 0.05~ 0.05~ 0.05            & 0.10~ 0.10~ 0.10            & 0.10~ 0.10~ 0.10           & 0.25~ 0.25~ 0.25           & 0.25~ 0.25~ 0.25                                \\
No Funneling               & 0.05~ 0.05~ 0.05             & 0.00~ 0.00~ 0.00            & 0.05~ 0.05~ 0.05            & 0.00~ 0.05~ 0.05           & 0.00~ 0.00~ 0.00           & 0.25~ 0.25~ 0.25                                \\
No Randomization           & 0.30~ 0.40~ 0.45             & 0.15~ 0.20~ 0.25            & 0.55~ 0.60~ 0.60            & 0.20~ 0.20~ 0.25           & 0.20~ 0.30~ 0.30           & 0.35~ 0.40~ 0.50                                \\
No Augmentation            & \textbf{0.50~ 0.75~ 0.80}    & 0.40~ 0.50~ 0.50            & 0.75~ 0.80~ 0.80            & \textbf{0.60~ 0.75~ 0.75}  & 0.60~ 0.75~ 0.75           & 0.65~ 0.75~ 0.75                                \\
\textbf{DexTransfer(ours)} & 0.40~ 0.55~ 0.70             & \textbf{0.50~ 0.70~ 0.80}   & \textbf{0.80~ 0.90~ 0.90}   & 0.55~ 0.75~ 0.75           & \textbf{0.80~ 0.80~ 0.85}  & \textbf{0.85~ 0.90~ 0.95}                       \\ 
\hline
                           & \uline{~ ~ ~ ~ mug~ ~ ~ ~ ~} & \uline{~bleach cleaner~}    & \uline{~ ~power drill~ ~}   & \uline{~ ~ large marker~~} & \uline{~ ~ pitcher base~~} & \uline{~ ~ ~~\textbf{\textbf{average}}~ ~ ~ ~}  \\
                           & ~ 1~ ~ ~2~ ~ ~3              & ~ 1~ ~ ~2~ ~ ~3             & ~ 1~ ~ ~2~ ~ ~3             & ~ 1~ ~ ~2~ ~ ~3            & ~ 1~ ~ ~2~ ~ ~3            & ~ 1~ ~ ~2~ ~ ~3                                 \\
Heuristic                  & 0.10~ 0.15~ 0.15             & 0.00~ 0.00~ 0.00            & \textbf{0.50~ 0.55~ 0.55}            & 0.10~ 0.10~ 0.10           & 0.00~ 0.00~ 0.00           & 0.11~ 0.12~ 0.13                                \\
Nearest Neighbor           & 0.15~ 0.15~ 0.15             & 0.05~ 0.05~ 0.05            & 0.10~ 0.10~ 0.10            & 0.15~ 0.20~ 0.20           & 0.00~ 0.00~ 0.00           & 0.14~ 0.15~ 0.15                                \\
No Funneling               & 0.05~ 0.25~ 0.30             & 0.00~ 0.05~ 0.05            & 0.00~ 0.00~ 0.05            & 0.00~ 0.00~ 0.00           & 0.00~ 0.00~ 0.00~          & 0.04~ 0.07~ 0.07                                \\
No Randomization           & 0.50~ 0.65~ 0.65             & 0.20~ 0.25~ 0.25            & 0.25~ 0.25~ 0.25            & 0.40~ 0.40~ 0.40           & 0.05~ 0.05~ 0.05           & 0.33~ 0.38~ 0.40                                \\
No Augmentation            & 0.50~ 0.65~ 0.70             & 0.10~ 0.20~ 0.25            & 0.15~ 0.20~ 0.25            & 0.90~ 0.90~ 0.90           & 0.05~ 0.05~ 0.05           & 0.47~ 0.55~ 0.57                                \\
\textbf{DexTransfer(ours)} & \textbf{0.85~ 0.85~ 0.85}    & \textbf{0.40 0.50~ 0.50}    & 0.40~ 0.45~ 0.45   & \textbf{0.90~ 0.90~ 0.90}  & \textbf{0.25~ 0.35~ 0.40}  & \textbf{0.61~ 0.70~ 0.74}                       \\
\hline
\end{tabular}
\end{table*}

In this section, we aim to understand which elements of our system are crucial to performance by evaluating the performance of the system as each component: domain randomization, augmentation and data funneling, is removed. In addition, we compare our approach with a heuristic based approach and a nearest neighbor approach. The results are shown in Table \ref{table:baselines} and are described below. 

\textbf{Policy without Domain Randomization}
We show that adding domain randomization in the dataset can significantly improve policy's ability to handle covariate shift at inference time. The key insight is one trajectory could be successful by chance, but cannot guarantee the robustness to noise and different physics parameters. We observe an almost $30\%$ performance drop without domain randomization when evaluating policies on 17 objects as show in the second row in Table \ref{table:baselines}. 

\textbf{Policy Without Augmentation}
We found without augmentation, the policy performs similar on smaller objects but worse on bigger objects like bleach cleaner, power drill and pitcher base. On average, the performance drops more than $10\%$. 

\textbf{Policy Without Data Funneling}
Without data funneling, the policy fails completely to grasp unseen poses from a novel initial robot state. This is because only refined trajectories cannot cover the entire space. We show that without data funneling, the success rate on average is less than $5\%$.

\textbf{Heuristic Approach}
We compare our approach with a heuristic approach, which drives the robot along a linear path to a certain distance from the center of the object pointclouds, and grasp the object using pre-defined finger poses. We observed that this policy only achieves an 11\% success rate. This shows that learning to adjust both position and orientation of the robot given the observation of the scene is crucial for success. 

\textbf{Nearest Neighbor}
To illustrate the difference between the training and test set, we further perform the following baseline: For each of the 20 test samples, we find the closest sample in the training set and execute its corresponding actions. We observe only a 14\% success rate on average. This suggests our policy learns to adapt to novel poses given the observation of the object. 

\subsection{Real Robot Experiments}
\label{sec:robot}
Lastly, we test policy transfer to the real world on a physical robot for a subset of the trained policies.

\begin{figure}[!ht]
    \centering
    \includegraphics[width=0.49\textwidth]{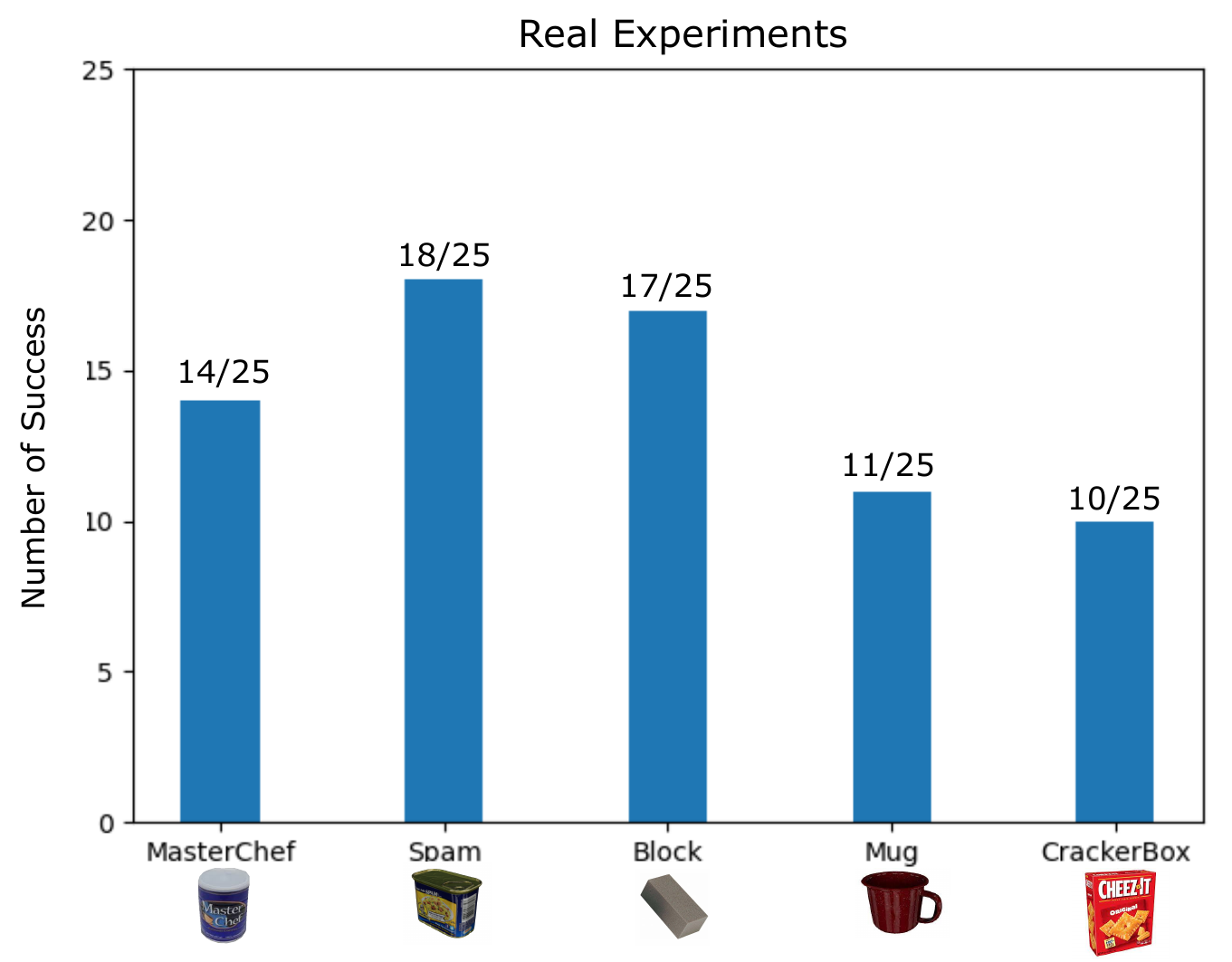}
    \caption{Real experiments with a 23-DoF Kuka Allegro robot tested on 5 objects. Each object is evaluated 25 times.}
    \label{fig:realE}
\end{figure}

\textbf{Robot System}
We deploy our policies to a robotic platform that has 23 actuators across a KUKA LBR
iiwa 7 R800 robot arm and a Wonik Robotics Allegro robotic hand. We place three RGB-D cameras around the robot to provide necessary point cloud information. Specifically, unseen object instance segmentation \cite{xiang:corl2020} is applied to the image data to extract a segmented point cloud of a single object on the table, which is used as input to the trained policies.

\textbf{Control}
The policies require additional control aspects for real-world application. First, the policy generates palm pose targets, which is an elevated action space for the robot. These pose targets are passed to an underlying, manually-derived geometric fabrics policy that generates high-frequency joint position, velocity, and acceleration targets motion across all joints of the arm. Geometric fabrics is a provably stable, second order policy that effectively solves the problem of reaching to end-effector targets while resolving arm redundancy, controlling manipulator posture, avoiding joint limits and joint speed limits, avoiding robot self-collision, and avoiding excessive collision between the robot and the table. The design and tuning of this policy is exactly the same as the one reported in \cite{van_wyk:ral2022}.

Finally, the target joint positions generated by fabrics are directly fed to an underlying gravity-compensated joint PD controller at 30 Hz, which are upsampled to 1000 Hz via polynomial interpolation. The trained policies also generate finger joint position commands at 6 Hz, which are fed directly to a PD controller that drives the joints of the Allegro hand. Altogether, the actions of the trained policies are ultimately converted to target joint drive torques that generate motion through the physical robot.

\textbf{Evaluation}
We tested policies on a subset of objects: Master chef can, potted meat can, mug, cracker box and wood block. In real world, wood block is too heavy for the Allegro hand to lift, so we replace it with a foam block with the same dimensions. To reduce the reflection of the object surface in order to get more reliable segmentation masks, we add non-sticky tapes around cracker box and potted meat can. We test our policies on each object 25 times and report success rate in Figure \ref{fig:realE}. Qualitative results are shown in Fig. \ref{fig:combined_qualitative}.  We see that while success rate in the real world is lower than in simulation, it is able to successfully complete the task reliably with over $40\%$ success rate for any object and in many cases closer to $70\%$ success rate.

\section{CONCLUSIONS}
In this work, we proposed a new system for learning to grasp various objects with a multi-fingered dexterous manipulator, leveraging a small number of human-provided demonstrations. Our system retargets the provided demonstrations into simulation, applies a technique for targeted data-augmentation and refinement and then applies large scale supervised learning to learn control policies that operate directly on point-cloud inputs. These policies when trained with sufficient augmentation can be transferred to the real world and can be used to grasp various objects with up to $72\%$ success rate. We show the efficacy of the system in both simulation and the real world, running careful analysis to understand the impact of various design decisions in the system. 

There are several directions for future work. For instance, it would be very interesting to go beyond tasks like grasping and do more dexterous problems like in-hand manipulation by continuing to leverage the same general system for simulation driven data augmentation. It would also be interesting to see if we can use the simulator to pretrain policies but continue to finetune them directly in the real world in new domains. Additionally, exploring more complex refinement algorithms like CMA-ES \cite{hansencmaes} rather than rejection sampling would be a worthwhile exercise.


\bibliographystyle{plainnat}
\bibliography{references}
\makeatletter
\g@addto@macro\@maketitle{
  \begin{figure}[H]
  \setlength{\linewidth}{\textwidth}
  \setlength{\hsize}{\textwidth}
  \centering
  \resizebox{0.98\textwidth}{!}{\includegraphics[]{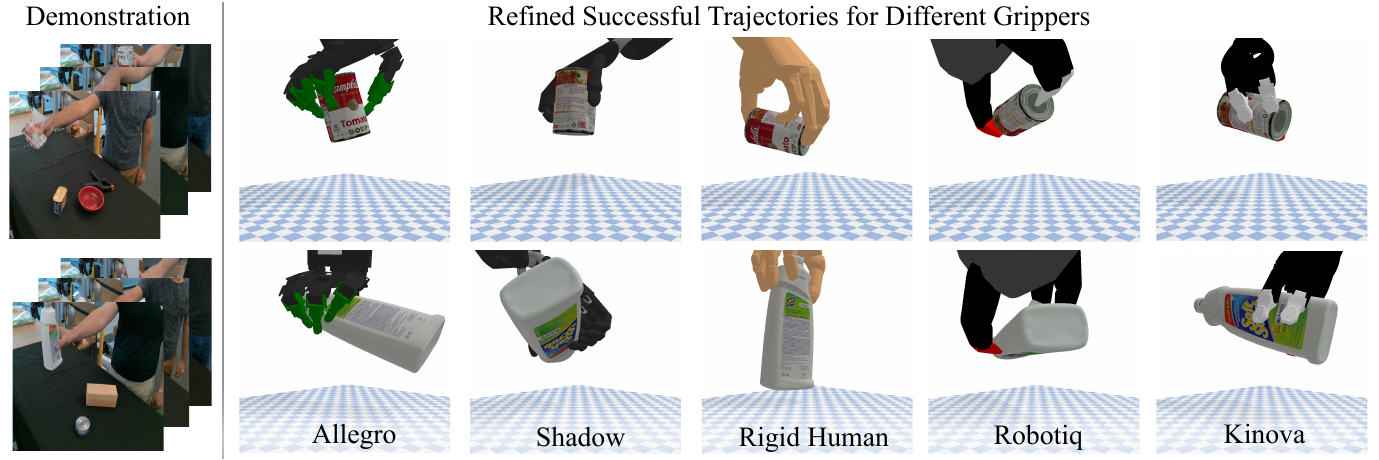}}
  \vspace{-2mm}
  \caption{Refined successful trajectories from human demonstrations to diverse robot grippers}
  \label{fig:diverse_gripper}
  \end{figure}
  \vspace{-4mm}
}

\appendices
\section{Related Work}

\paragraph{\textbf{Manipulation with Dexterous Hands}}~Dexterous manipulation has a long history dating back to~\cite{salisbury:ijrr1982,mason:1985}. More recently, the field has studied the use of dexterous hands in a variety of task domains, ranging from object grasping~\cite{mandikal:icra2021,mandikal:corl2021}, in-hand manipulation~\cite{bai:icra2014,kumar:icra2016,openai:arxiv2018,openai:arxiv2019,chen:corl2021}, relocation~\cite{kumar:icra2014,qin:arxiv2021}, stacking~\cite{jeong:corl2020}, interaction with environmental props~\cite{gupta:iros2016,zhu:icra2019}, tool use~\cite{rajeswaran:rss2018,jain:icra2019,nagabandi:corl2019,radosavovic:iros2021}, to smart teleoperation~\cite{handa:icra2020,garcia-hernando:iros2020}. Our work addresses the task of object grasping, a key first step to gain hold of an object before any downstream manipulation tasks. Grasping has been conventionally approached by model-based planning~\cite{miller:ram2004}. Notably, for simpler end effectors like parallel-jaw grippers, remarkable progress has been recently achieved through end-to-end learning-based approaches which directly predict grasp position from raw visual input, such as a depth image~\cite{mahler:rss2017} or point cloud~\cite{mousavian:iccv2019}. Our work is in line with this direction but addresses the frontier of the more complex dexterous hands.

To control a dexterous hand, prior work has also investigated different approaches, including planning with analytical models~\cite{bai:icra2014} and online trajectory optimization~\cite{kumar:icra2014}. However, these methods assume accurate dynamics models and robust state estimates, which are difficult to obtain in complex real-world manipulation. To overcome this limitation, recent work has resorted to learning-based approaches particularly with deep reinforcement learning (RL). Both model-based RL~\cite{kumar:icra2016,gupta:iros2016,nagabandi:corl2019} and model-free RL~\cite{openai:arxiv2018,openai:arxiv2019,jeong:corl2020,chen:corl2021} have been investigated. Despite the progress, training deep RL models remains notoriously challenging due to high sample complexity and tedious reward engineering. Although this issue has been mitigated by incorporating human demonstrations~\cite{rajeswaran:rss2018,zhu:icra2019,qin:arxiv2021,radosavovic:iros2021,mandikal:corl2021}, these methods are still faced with a major challenge in scalability---training a single general policy for handling diverse objects and diverse scene configurations is still largely beyond reach~\cite{qin:arxiv2021,chen:corl2021}. Rather than using deep RL, we adopt the common supervised learning paradigm, where we focus on bootstrapping a large dexterous grasping dataset from a small set of human demonstrations. Our work bears some similarity to~\cite{jain:icra2019,chen:corl2021}, where an expert policy is first trained using RL in a privileged state space, followed by training a high-dimensional student policy from expert generated data via behavioral cloning. Instead of training RL experts, our method bootstraps from minimal human demonstrations. Besides, as opposed to many prior works which only evaluate in simulation~\cite{rajeswaran:rss2018,jain:icra2019,jeong:corl2020,mandikal:icra2021,radosavovic:iros2021,mandikal:corl2021,chen:corl2021}, we evaluate our model on a real-world robot platform.
\begin{figure}[h]
    \centering
    \includegraphics[width=0.49\textwidth]{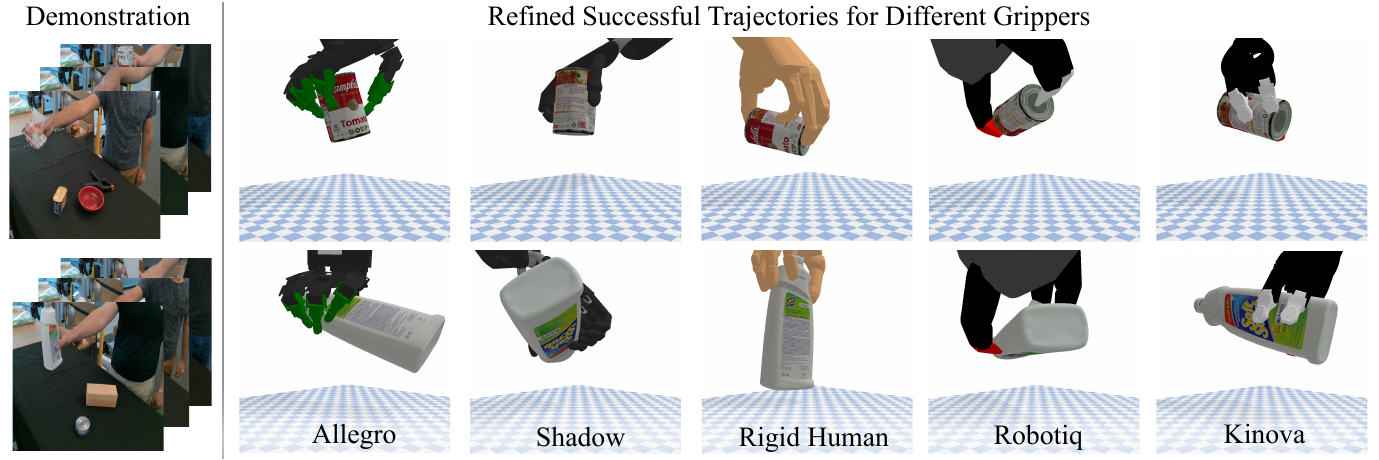}
    \caption{Refined successful trajectories from human demonstrations to diverse robot grippers}
    \label{fig:diverse_gripper}
\end{figure}
\begin{figure*}[h]
    \centering
    \includegraphics[width=\textwidth]{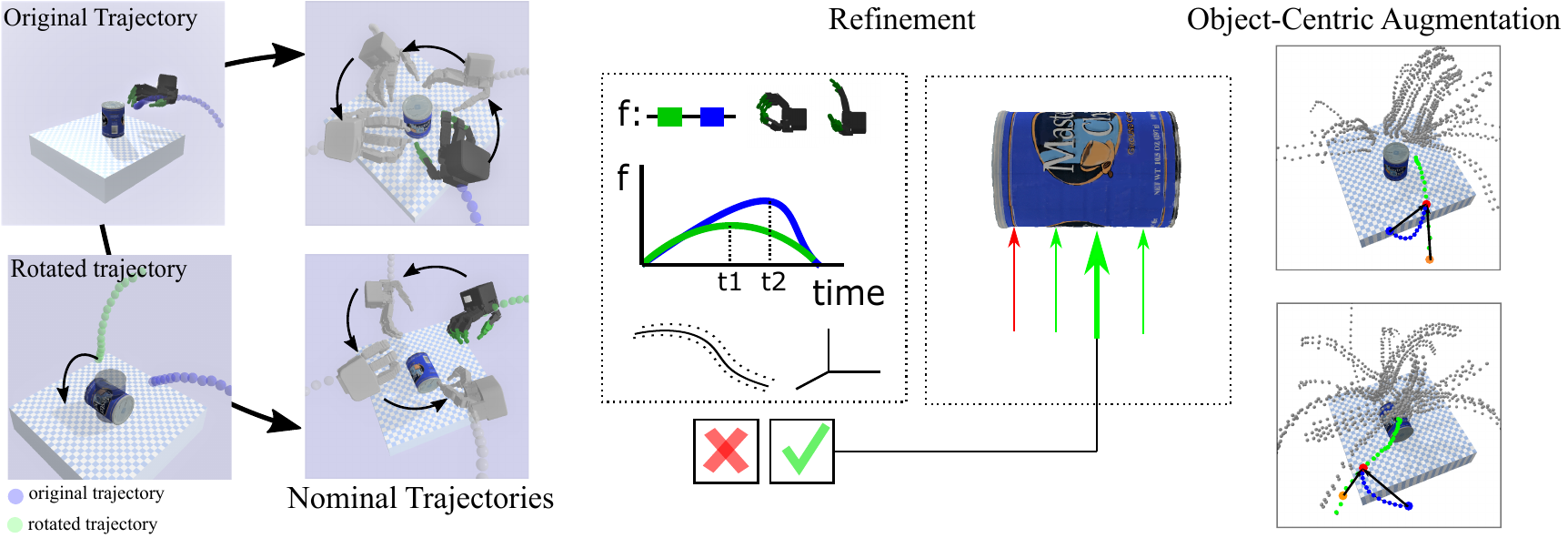}
    \caption{Illustration of trajectory refinement procedure combining refinement, augmentation, and data funneling to get an extended dataset of trajectories for supervised learning. \textbf{Left:} Only a few human demonstrations are provided and retargeted to a robot gripper to generate nominal trajectories. \textbf{Middle:} grasping trajectories generated by policies on unseen object poses and initial hand poses via refinement and augmentation. \textbf{Right:} Extending the set of feasible object poses and novel hand states via data funneling.}
    \label{fig:trajectory_refinement}
\end{figure*}
\begin{table*}
\centering
\caption{Experiments Illustrating Effect of Occluded Pointclouds on Policy Learning}
\label{table:ablation}
\begin{tabular}{|P{3cm}|P{2cm}|P{2cm}|P{2cm}|P{2cm}|P{2cm}|P{2cm}|} 
\hline
\begin{tabular}[c]{@{}l@{}}\\\\\underline{}\\\end{tabular} & \underline{master chef can }                             & \underline{~ ~cracker box~ ~}                            & \underline{~ ~ ~sugar box~ ~ ~}                          & \underline{tomato soup can }                             & \underline{~ mustard bottle~}                            & \underline{~ ~tuna fish can~~}                            \\
                                                       & ~ 1~ ~ ~2~ ~ ~3                                    & ~  1~ ~ ~2~ ~ ~3                                    & ~ 1~ ~ ~2~ ~ ~3                                    & ~ 1~ ~ ~2~ ~ ~3                                    & ~ 1~ ~ ~2~ ~ ~3                                    & ~ 1~ ~ ~2~ ~ ~3                                     \\
\textcolor[rgb]{0.251,0.231,0.243}{vertices}           & \textcolor[rgb]{0.251,0.231,0.243}{0.90~ 0.90~ 0.90} & \textcolor[rgb]{0.251,0.231,0.243}{0.45~ 0.50~ 0.50} & \textcolor[rgb]{0.251,0.231,0.243}{0.70~ 0.85~ 0.85} & \textcolor[rgb]{0.251,0.231,0.243}{0.80~ 0.85~ 0.85} & \textcolor[rgb]{0.251,0.231,0.243}{0.55~ 0.65~ 0.65} & \textcolor[rgb]{0.251,0.231,0.243}{0.60~ 0.75~ 0.80}  \\
1 cam                                                  & 0.25~ 0.30~ 0.30                                     & 0.15~ 0.15~ 0.15                                     & 0.35~ 0.40~ 0.40                                     & 0.25~ 0.25~ 0.25                                     & 0.25~ 0.25~ 0.25                                      & 0.15~ 0.15~ 0.20                                      \\
4 cams                          & \textbf{0.70~ 0.80~ 0.80}                            & 0.20~ 0.30~ 0.35                                     & \textbf{0.65~ 0.70~ 0.80}                            & \textbf{0.70~ 0.85~ 0.85}                            & \textbf{0.75~ 0.85~ 0.85 }                           & 0.70~ 0.80~ 0.85                                      \\
4 cams+contact                                         & 0.60~ 0.60~ 0.60                                     & \textbf{0.25~ 0.35~ 0.35}                            & 0.55~ 0.65~ 0.75                                     & 0.65~ 0.75~ 0.80                                     & 0.75~ 0.75~ 0.75                                     & \textbf{0.90~ 1.00~ 1.00}                             \\
\hline
                                                       & \underline{~ ~pudding box~ ~}                            & \underline{~ ~ gelatin box~ ~~}                          & \underline{potted meat can}                              & \underline{~ ~wood block~~}                              & \underline{~ ~foam brick~~}                              & \underline{~ ~ ~ ~bowl~ ~ ~ ~}                        \\
                                                       & ~ 1~ ~ ~2~ ~ ~3                                    & ~ 1~ ~ ~2~ ~ ~3                                    & ~ 1~ ~ ~2~ ~ ~3                                    & ~ 1~ ~ ~2~ ~ ~3                                    & ~ 1~ ~ ~2~ ~ ~3                                    & ~ 1~ ~ ~2~ ~ ~3                                     \\
\textcolor[rgb]{0.251,0.231,0.243}{vertices}           & \textcolor[rgb]{0.251,0.231,0.243}{0.45~ 0.75~ 0.75} & \textcolor[rgb]{0.251,0.231,0.243}{0.50~ 0.60~ 0.75} & \textcolor[rgb]{0.251,0.231,0.243}{0.80~ 0.85~ 0.90} & \textcolor[rgb]{0.251,0.231,0.243}{0.90~ 0.90~ 0.90} & \textcolor[rgb]{0.251,0.231,0.243}{0.85~ 0.95~ 0.95} & \textcolor[rgb]{0.251,0.231,0.243}{0.70~ 0.95~ 0.95}  \\
1 cam                                                  & 0.15~ 0.15~ 0.15                                     & 0.10~ 0.10~ 0.20                                     & 0.25~ 0.30~ 0.30                                     & 0.35~ 0.35~ 0.35                                     & 0.35~ 0.35~ 0.35                                     & 0.35~ 0.35~ 0.35                                      \\
4 cams                                                 & 0.40~ 0.55~ 0.70                                     & 0.50~ 0.70~ 0.80                                     & \textbf{0.80~ 0.90~ 0.90}                            & 0.55~ 0.75~ 0.75                                     & \textbf{0.80~ 0.80~ 0.85}                            & 0.80~ 0.95~ 0.95                                      \\
4 cams+contact                                         & \textbf{0.55~ 0.60~ 0.65}                            & \textbf{0.75~ 0.75~ 0.80}                            & 0.70~ 0.85~ 0.85                                     & \textbf{0.60~ 0.80~ 0.80}                            & 0.75~ 0.80~ 0.85                                     & \textbf{0.90~ 1.00~ 1.00}              \\ 
\hline
                                                       & \underline{~ ~ ~ ~ ~mug~ ~ ~ ~}                        & \underline{~bleach cleaner~}                             & \underline{~ ~power drill~ ~}                            & \underline{~ ~ large marker~~}                           & \underline{~ ~ pitcher base~~}                           & \underline{~ ~ ~ \textbf{average}~ ~ ~ ~}                 \\
                                                       & ~ 1~ ~ ~2~ ~ ~3                                    & ~ 1~ ~ ~2~ ~ ~3                                    & ~ 1~ ~ ~2~ ~ ~3                                    & ~ 1~ ~ ~2~ ~ ~3                                    & ~ 1~ ~ ~2~ ~ ~3                                    & ~ 1~ ~ ~2~ ~ ~3                                     \\
\textcolor[rgb]{0.251,0.231,0.243}{vertices}           & \textcolor[rgb]{0.251,0.231,0.243}{0.75~ 0.80~ 0.80} & \textcolor[rgb]{0.251,0.231,0.243}{0.60~ 0.70~ 0.70} & \textcolor[rgb]{0.251,0.231,0.243}{0.45~ 0.55~ 0.55} & \textcolor[rgb]{0.251,0.231,0.243}{0.85~ 0.85~ 0.85} & \textcolor[rgb]{0.251,0.231,0.243}{0.25~ 0.40~ 0.55} & \textcolor[rgb]{0.251,0.231,0.243}{0.65~ 0.75~ 0.79}  \\
1 cam                                                  & 0.35~ 0.40~ 0.40                                     & 0.35~ 0.35~ 0.35                                     & 0.15~ 0.15~ 0.15                                     & 0.50~ 0.50~ 0.50                                       & 0.05~ 0.05~ 0.05                                     & 0.25~ 0.26~ 0.27                                      \\
4 cams                                                 & 0.85~ 0.85~ 0.85                                     & 0.40~ 0.45~ 0.55                                     & 0.40~ 0.45~ 0.45                                     & 0.90~ 0.90~ 0.90                                      & 0.25~ 0.35~ 0.40                                     & 0.61~ 0.70~ 0.74                                      \\
4 cams+contact                                         & \textbf{0.90~ 0.90~ 0.90}                            & \textbf{0.40~ 0.60~ 0.60}                            & \textbf{0.50~ 0.50~ 0.55}                            & \textbf{0.95~ 0.95~ 0.95}                            & \textbf{0.35~ 0.35~ 0.35}                            & \textbf{0.65~ 0.72~ 0.74}                             \\
\hline
\end{tabular}
\end{table*}

\paragraph{\textbf{Learning from Human Demonstration}}~While human demonstration plays a crucial role in training manipulation policies, devising a scalable means to collect demonstrations is still a key bottleneck. Prior work has collected demonstrations kinesthetically~\cite{zhu:icra2019}, through VR interfaces~\cite{rajeswaran:rss2018}, or using motion capture (mocap) solutions~\cite{garcia-hernando:cvpr2018,hasson:cvpr2019,taheri:eccv2020,qin:arxiv2021}. Although there has been some innovations for parallel-jaw grippers~\cite{kokic:ral2020,song:ral2020,young:corl2020}, acquiring demonstrations for dexterous manipulators remains challenging. Consequently, the collected demonstration datasets are often limited in size. Our work leverages existing mocap datasets of human grasping~\cite{chao:cvpr2021} to guide the generation of a large, diverse grasping dataset for training dexterous manipulators.
Apart from our work, some recent efforts have also explored collecting demonstrations by harvesting internet video~\cite{mandikal:corl2021}.

\paragraph{\textbf{Multi-Finger Grasp Synthesis}}~Our work connects closely to the task of multi-finger grasp synthesis. Prior work has investigated grasp synthesis for virtual human hands~\cite{taheri:eccv2020,karunratanakul:3dv2020} and multi-finger robot manipulators~\cite{brahmbhatt:iros2019}. However, they are only concerned with static grasp poses and do not consider the motion. Some very recent work attempted to bridge this gap by synthesizing trajectories for human-like hands~\cite{zhang:siggraph2021,christen:arxiv2021}. Our work can potentially capitalize on their grasp trajectories by utilizing them as additional human demonstrations.

\section{DexTransfer}

\subsection{Generalization on Different Grippers}
We further investigate whether DexTransfer is general for other grippers. In addition to Allegro robot, we also generate a dataset for Shadow hand, MANO hand, Robotiq-3f gripper and Kinova. Fig\ref{fig:diverse_gripper} shows qualitative results on refined poses from human demonstrations. This suggests that DexTransfer is able to transfer human demonstrations to various robot grippers despite the large difference among their kinematic configurations. 
\begin{figure*}[h]
\setlength{\linewidth}{\textwidth}
\setlength{\hsize}{\textwidth}
\centering
\resizebox{0.98\textwidth}{!}{\includegraphics[]{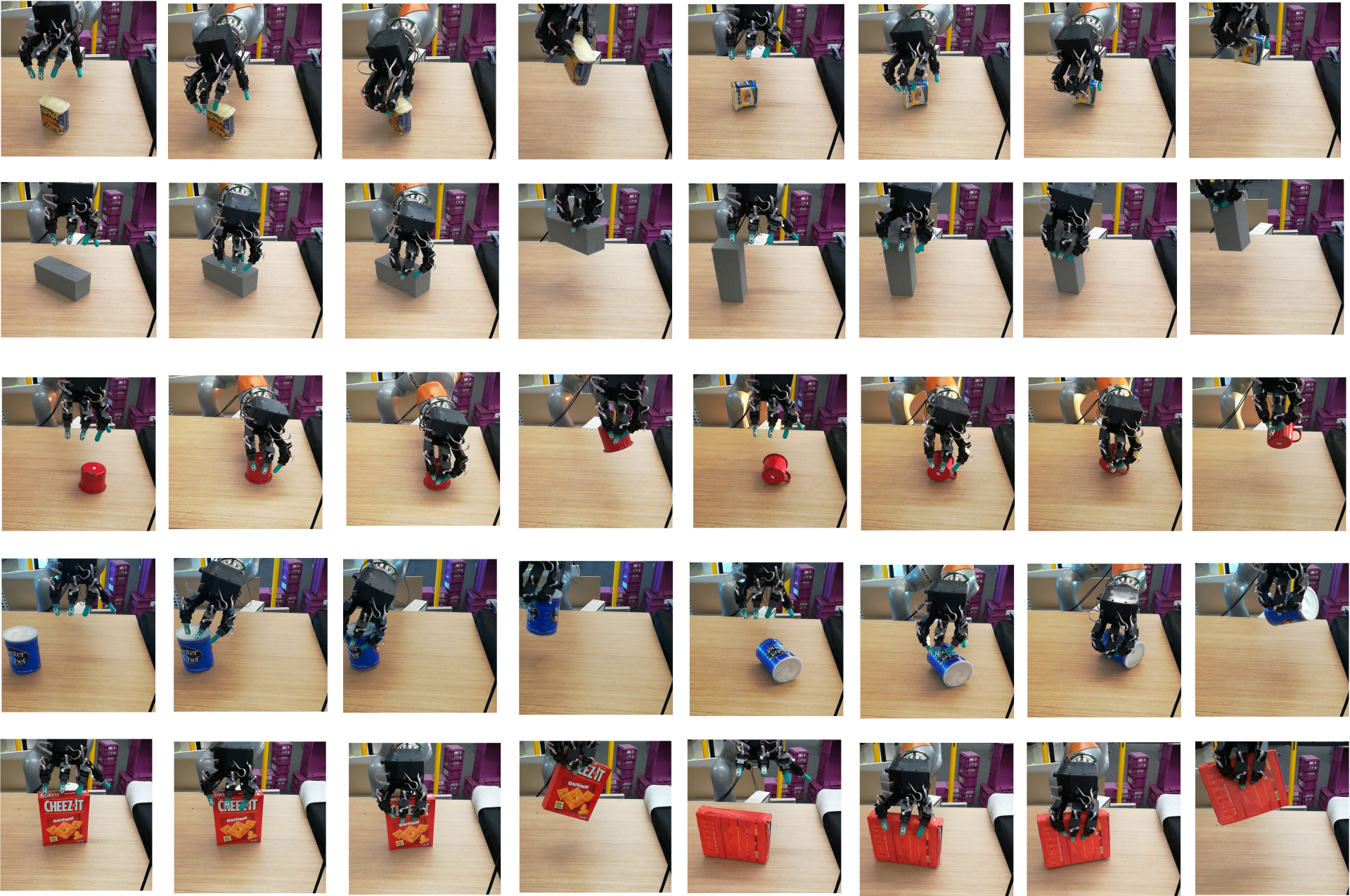}}
\vspace{-2mm}
\caption{Examples of policies on unseen poses in real world}
\label{fig:qualitative_real}
\end{figure*}
\begin{figure}[h]
    \centering
    \includegraphics[width=0.48\textwidth]{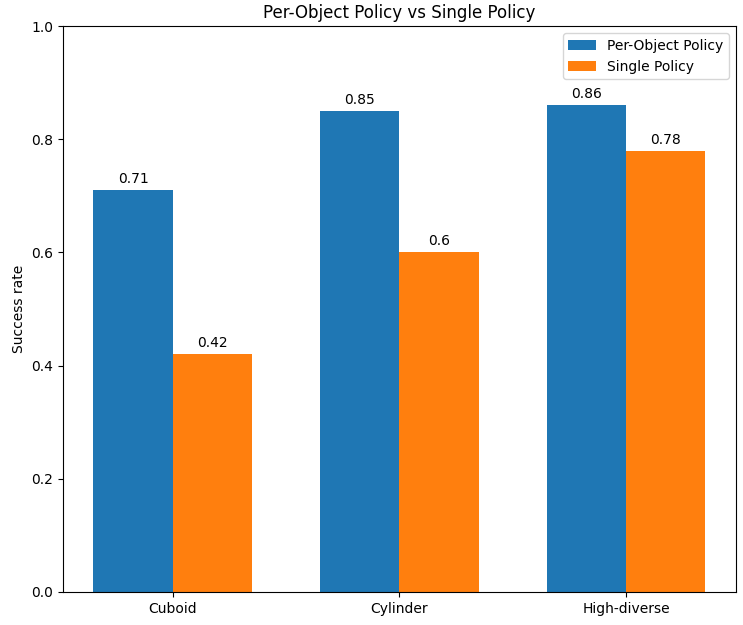}
    \caption{Single policy across multiple objects}
    \label{fig:single_policy}
\end{figure}

\subsection{Details on Trajectory Refinement}
Fig \ref{fig:trajectory_refinement} shows details on trajectory refinement and augmentation in Dextransfer. For each nominal trajectory,  we sample a scalar parameter time $t$ from the uniform distribution $t \sim \mathcal{U}[0, 1]$, maximum palm pose $(p_{\text{max}}^i)_{k}$ and minimum palm pose $(p_{\text{min}}^i)_{k}$. We then use this parameter to interpolate between minimum and maximum perturbation values within the corresponding time steps $(p_{t}^i)_{k} = t*(p_{\text{max}}^i)_{k} + (1-t)*(p_{\text{min}}^i)_{k}$, which can then be added to $a_{t}^i$ to obtained the perturbed action. We add domain randomization and only keep refined actions that are successful with 10 different physics parameters (object mass, frictions, noise and perturbation test) via rejection sampling. 
\section{Experiment}
\subsection{Dataset Generation Details}
In real world experiments, we found Allegro hand is not strong enough to lift heavy objects like power drill or bleach cleaner, so we reduce all the object mass to less than $0.25\ kg$. During data randomization, we fixed Allegro hand torque force that is close to a real robot: $0.6\ N.m$, and randomize object mass $m \in \left[0.1\ kg, 0.25\ kg\right]$, and object friction coefficient $\mu \in \left[0.7, 0.85\right]$. We injected noise $\sim \mathcal{N}(0,\, 0.01)m$ to each candidate trajectory. In addition, we add $3$ seconds of high-frequency perturbation to the palm at the end of the grasping trajectory to further identify unstable grasping poses. We use PyBullet \cite{coumans:2021} to generate the dataset and the robot is controlled at 12Hz in simulation. 

\subsection{Ablation on Input Representation}
\begin{figure*}[h]
    \centering
    \includegraphics[width=\textwidth]{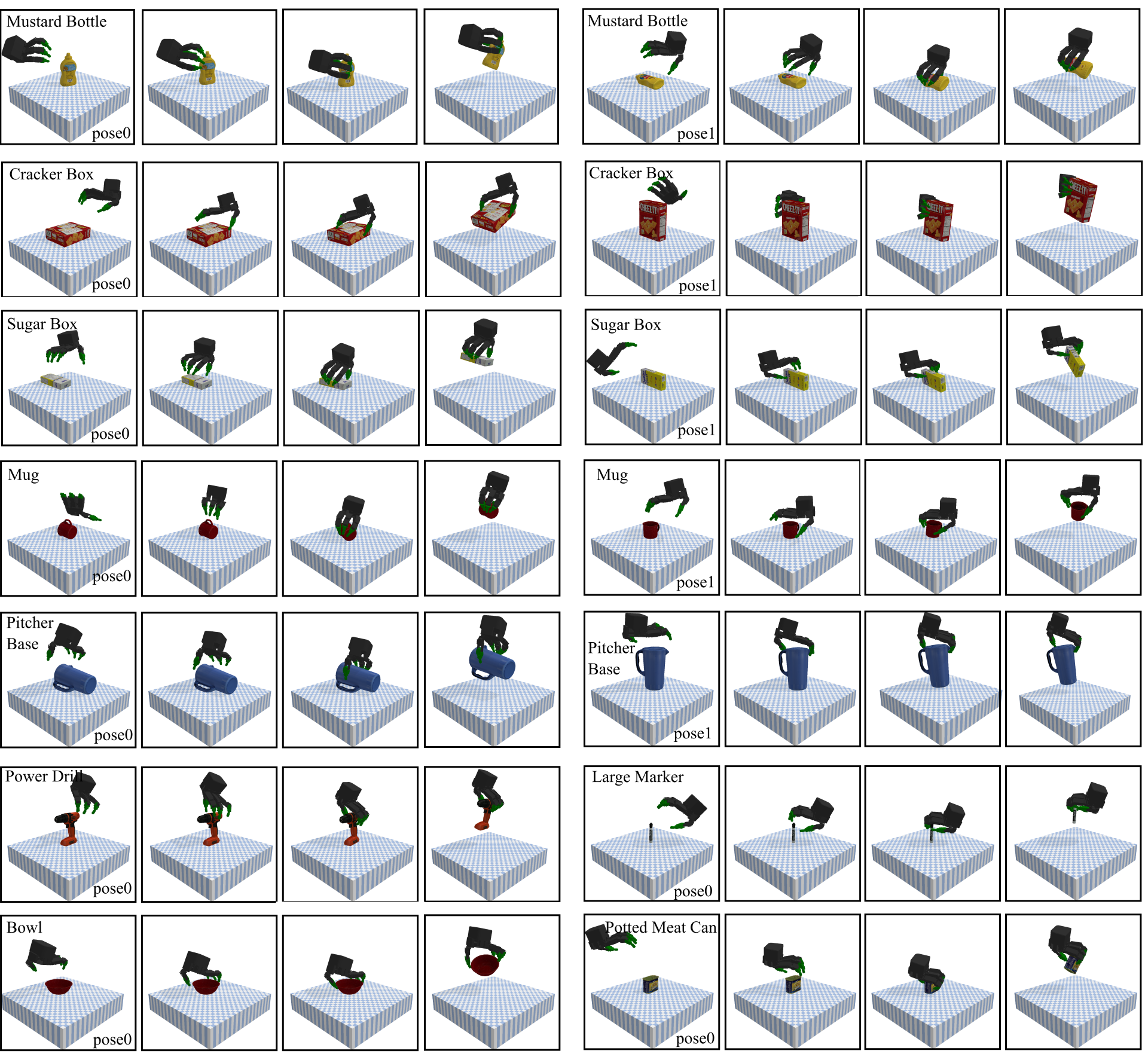}
    \caption{Examples of policies on unseen poses in simulation}
    \label{fig:qualitative_sim}
\end{figure*}
\label{sec:ablation}
We aim to understand the impact that learning on noisy or occluded point clouds has on policy learning. In particular, we consider the impact of acquiring the point cloud using just a single camera versus using multiple cameras, as well as including the ability to sense binary contacts to deal with partial observability. In particular, we compare the following scenarios for pointcloud acquisition (as detailed in Table \ref{table:ablation}):

\textbf{Full Vertices (Oracle):} This assumes no occlusion during robot movement, and the state is fully observable. Note that this will not be realistic for any real robot experiments, but its useful as a upper-bound on the performance of the partially observable case. On average, it achieves $65\% $, $75\%$ and $79\%$ success rate after three successive trials. 

\textbf{One Camera:} If we only use one camera to render the scene, severe occlusion occurs when the robot is close to the object. We randomize camera position during training. We found that the policy performs poorly in this scenario, on average, only $25\%$, $26\%$ and $27\%$ success rate are achieved. 

\textbf{Four Cameras:} Two cameras are placed on top of the object, and two are on the side. We found that with 4 cameras, the policy is able to achieve similar performance as sampling from vertices: $61\%$, $70\%$ and $74\%$.

\textbf{Binary Contact:}
Occlusion is inevitable during real world experiments, we investigate whether adding extra information about per-finger binary contact could help the policy under partially observable conditions. For each finger tip link, we set a binary contact variable to 1---if the contact force is larger than 0.5N---and 0 otherwise. We concatenate this binary contact feature with robot poses as the input for the Kinematics Encoder. We found that this extra information can help to fill the gap in performance between 4 camera rendering and full vertices. In particular, it helps to improve performance on smaller objects, for example, tuna fish can, pudding box, gelatin box, and large maker, where occlusion usually occurs the most as shown in Table \ref{table:ablation}. Interestingly, we found it performs slightly worse on some objects, for example master chef can and sugar box. We hypothesize the reason is that binary contact sometimes might mislead the network to think the grasp is stable enough to lift, and that a richer non-binary contact format could help with this issue.

\subsection{Single Policy across Multiple Objects}
We also show that we can train a single policy that can perform grasping across multiple objects in Figure\ref{fig:single_policy}. We group objects with similar shapes: Cuboids and Cylinders. Each group contains 5 objects. Cuboid Objects include cracker box, sugar box, gelatin box, wood block and foam brick. Cylindrical objects include master chef can, tomato soup can, tuna fish can, mug and large marker. In addition, we select 5 objects that have diverse shapes. In order to evaluate the dropping performance when training across multiple diverse objects, we select 5 objects that yielded high performance when training per-object policies: master chef can, potted meat can, bowl, mug and large marker. We compute the average success rate from the per-object policies as an upper bound, and compare how the performance drops when training a single combined policy.

\subsection{Qualitative Results}
We show more qualitative results in both simulation (Fig.\ref{fig:qualitative_sim}) and real robot experiments (Fig.\ref{fig:qualitative_real}) . Our policy learns to adjust poses and fingers smoothly while approaching to the object. Further results can be found on the supplementary website \mbox{\url{https://sites.google.com/view/dextransfer/home}}

\end{document}